\documentclass{article}

     \PassOptionsToPackage{numbers, compress}{natbib}

\usepackage[final]{neurips_2022}

\usepackage[utf8]{inputenc} 
\usepackage[T1]{fontenc}    
\usepackage{hyperref}       
\usepackage{url}            
\usepackage{booktabs}       
\usepackage{amsfonts}       
\usepackage{nicefrac}       
\usepackage{microtype}      
\usepackage[dvipsnames]{xcolor}         
\usepackage{bm}
\usepackage{wrapfig}
\usepackage{graphicx}
\usepackage{amsmath}
\usepackage{cleveref}
\usepackage{bbm}

\usepackage{caption}
\usepackage{subcaption}
\usepackage{booktabs}
\usepackage{array,multirow}

\newcommand{\bo}[1]{\textbf{#1}}
\newcommand{\bw}[1]{\textit{#1}}

\newcommand\floor[1]{\lfloor#1\rfloor}
\newcommand\ceil[1]{\lceil#1\rceil}

\title{Neural Combinatorial Logic Circuit Synthesis from Input-Output Examples}

\author{
Peter Belcak \\
ETH Z\"urich\\
\texttt{belcak@ethz.ch}
\And
Roger Wattenhofer \\
ETH Z\"urich\\
\texttt{wattenhofer@ethz.ch}
}

\begin{document}

\maketitle

\begin{abstract}
We propose a novel, fully explainable neural approach to synthesis of combinatorial logic circuits from input-output examples.
The carrying advantage of our method\footnote{We make all our code and data available at \url{https://github.com/pbelcak/neccs}.} is that it readily extends to \textit{inductive} scenarios, where the set of examples is incomplete but still indicative of the desired behaviour.
Our method can be employed for a virtually arbitrary choice of atoms -- from logic gates to FPGA blocks -- as long as they can be formulated in a differentiable fashion, and consistently yields good results for synthesis of practical circuits of increasing size.
In particular, we succeed in learning a number of arithmetic, bitwise, and signal-routing operations, and even generalise towards the correct behaviour in inductive scenarios.
Our method, attacking a discrete logical synthesis problem with an explainable neural approach, hints at a wider promise for synthesis and reasoning-related tasks.
\end{abstract}

\section{Introduction}
\label{section:introduction}

Logic circuit synthesis is the process of producing a logic circuit that satisfies a given specification and is a classical problem in computer science.
The synthesis of high-level designs to circuits is typically done as direct compilation of hardware description language code coupled with post-processing optimisation \cite{barbacci1975comparison,reilly2003milestones}, although recent work hints at that there exists room for merging the two by the use of neural compiler architectures \cite{roziere2020unsupervised}.
Transformer-based neural architectures have also been used for comprehension of linear temporal logic \cite{hahn2020teaching}, its synthesis (as the input, high-level specification) into circuits \cite{schmitt2021neural}, and the understanding of abstract mathematical patterns more generally \cite{belcak2022fact}.

Synthesising logic circuits from input-output listings, on the other hand, falls under the paradigm of programming by example.
If the listings are incomplete (i.e. not all possible inputs are considered) but indicative of the desired behaviour, one talks of \textit{circuit induction}.
Note that a well-posed instance of circuit induction still has a unique desired behaviour for the target circuit but is nevertheless more difficult, since any attempt to address it has to incorporate the guiding assumptions while relying on example data that may potentially contradict some interpretations of these assumptions.

We focus on the synthesis of combinatorial logic circuits from input-output examples.
This old problem of theoretical interest to mathematical logic \cite{church1963application,buchi1990solving} has recently been enjoying a small renaissance due to plethora of specific machine-learning applications \cite{boroumand2021learning,chowdhury2021openabc,haaswijk2018deep,hosny2020drills,rokach2012machine,rai2021logic}.
The traditional synthesise-and-optimise approaches \cite{hassoun2001logic} do not scale well and and are not applicable for later deployment on modern, by design already heavily optimised hardware.
Further, it is often far from practical to describe the desired circuit in terms of input-output examples \textit{completely}, and there is therefore an appetite for methods that could induce the correct behaviour from a few guiding examples, or query the user interactively.

To this end, we present a new method for synthesising circuits for combinatorial behaviour utilising any appropriately formulated design of atomic unit.
It is in the design of this unit where the user may choose to incorporate their implicit inductive assumptions, say in the form of making certain operations readily available (so that they need not be learned) or initialising decision weights in favour of particular configurations.
Our method relies on a neural-network-like layout of atomic units and admits training with gradient descent and loss, as is common in deep learning.
In situations where some examples are missing from the input-output specification, our method manages to incorporate the implicit design bias of the atoms to generalise well to previously unseen inputs, which is in its own right a challenging problem for classical deep neural networks \cite{belcak2022periodic}.

As a broader message of this workshop contribution, we wish to suggest that with appropriate architectures and inductive biases in place, neural methods can feasibly be used to tackle traditionally strictly discrete and reasoning-heavy problems such as circuit synthesis while achieving high levels of accuracy and possessing full model explainability through its outputs.
We further believe that neural methods may be the natural choice for situations where inductive assumptions need to be incorporated into the synthesis and reasoning processes.

\textbf{Formal setting.} Let $f: \left\{0,1\right\}^{w_i} \to \left\{0,1\right\}^{w_o}$ be a Boolean function and let $\mathcal{D} := \left\{ \left(x, f\left(x\right)\right) : x \in \left\{0,1\right\}^{w_i} \right\}$ be the dataset of all input-output pairs.
We call the task of constructing a circuit $\mathcal{C}$ from $\mathcal{D}$ such that $\mathcal{C}(x) = f(x) \,\forall x \in \left\{0,1\right\}^{w_i}$ \textit{circuit synthesis} from (complete) examples.
If, instead of $\mathcal{D}$, we are given $\mathcal{D}' \subset \mathcal{D}$, constructing $\mathcal{C}$ from $\mathcal{D}'$ is called circuit induction.


\section{Method}
\label{section:method}
Our method can be used to address both the task of circuit synthesis and induction.
It encompasses the translation of discrete logical units such as traditional logic gates or FPGA blocks, and the feed-forward combinatorial differentiable wiring technique that allows connections to emerge based on need.

Given these logical units, the challenge of the traditional circuit synthesis is to make an appropriate choice of available units and then wire them in a fashion that leads to correct functionality.
In contrast, our method relies on a single choice of the universal unit type that is made before the synthesis.

\subsection{Logical Units from Differentiable Operators}
For given differentiable input signal lines $i_1,i_2 \in \left[0, 1\right] \subset \mathbb{R}$, the basic logical operations such as NOT, AND, OR, or XOR can be readily translated to their differentiable counterparts as $1-i_1$, $i_1 i_2$, $i_1+i_2$, $i_1(1-i_2) + i_2(1-i_1)$, respectively.
As any Boolean function $f: \left\{0,1\right\}^{w_i} \to \left\{0,1\right\}^{w_o}$ can be represented as a network of a subset of these gates, such $f$ may also be differentiably implemented simply by composing the individual gates' differentiable counterparts appropriately.
Thus, any Boolean function representing a combinatorial circuit can be implemented with the use of differentiable operators.

Unlike the traditional synthesis setting where the algorithm has to make the choices of \textit{which unit to use} and \textit{what to connect it with} to ensure correctness, we decide to use only units that are \textit{universal} in the sense that a homogeneous network of such properly configured units can represent any $f$ as above.
Note that our approach to network building (cf. \Cref{section:softax-choice_wiring}) extends to the scenario where multiple types of units are to be interconnected, but such situations raise the problem of arrangement of heterogeneous units into network layers so that an arbitrary function can still be learned. 

We consider three chosen types for the universal unit, namely the and-inverter gate (AIG), 4-to-1 look-up table (LUT), and the corresponding 4-to-1 LUT-adder block (LAB).

\textbf{And-inverter gate.} The (2-bit) AIG simply computes $1-i_1 i_2$ for input signals $i_1,i_2$ as above.

\textbf{Look-up table.} Our LUT is a continuous variant of look-up tables commonly found in field-programmable gate arrays \cite{hassoun2001logic} and consists of a 16-dimensional learned (bias) tensor $\mathcal{T}$ arranged into four axes corresponds to a filtering decision on the four inputs to compose the final output. Let $\bm{i}$ be the vector of inputs, denote $\mathcal{I}$ a list of indices, $0 \leq \left|\mathcal{I}\right|$ its length, and $\epsilon$ the empty list.
Then the output of the lookup operation $\text{LUT}\left(\bm{i}\right)$ can be computed as
\begin{align*}
    \text{LUT}\left(\bm{i}\right) & := \text{LUT}\left(\bm{i}, \epsilon\right) \\
    \text{LUT}\left(\bm{i}, \mathcal{I}\right) & := \bm{i}_{1 + \left|\mathcal{I}\right|} \text{LUT}\left(\bm{i}, \mathcal{I}{:}1\right) + \left(1 - \bm{i}_{1 + \left|\mathcal{I}\right|}\right) \text{LUT}\left(\bm{i}, \mathcal{I}{:}0\right) & \text{for $\left|\mathcal{I}\right| < 4$} \\
    \text{LUT}\left(\bm{i}, \mathcal{I}\right) & := \mathcal{T}\left[\mathcal{I}\right] & \text{for $\left|\mathcal{I}\right| = 4$} 
\end{align*}
where $\mathcal{I}{:}n$ denotes the list formed by appending $n$ to $\mathcal{I}$ and $\mathcal{T}\left[\mathcal{I}\right]$ denotes the indexing the four-axis table tensor $\mathcal{T}$ by the four indices (deterministic; either $0$ or $1$) listed by $\mathcal{I}$ in order.

\textbf{LUT-adder block.}

The LAB is a LUT combined with a 1-bit full adder, and its output is decided as a learned softmax choice (constant attention) over the two outputs of the adder (including carry) and the one output of the LUT. 
The adder, which is learnably an option for the output of the LAB unit, is an example of an architectural inductive bias towards addition.
Let $\bm{c}$ be the learned parameter vector for the constant attention $\bm{\alpha}$, $\bm{i}$ vector of inputs as above, and let $\bm{\alpha} := \textit{softmax}\left( \bm{c} \right)$.
Then
\[
    \text{LAB}\left( \bm{i} \right) := \bm{\alpha}_1 \left( \Sigma \bmod 2 \right) + \bm{\alpha}_2 \mathbbm{1}_{\Sigma \geq 2} + \bm{\alpha}_3 \text{LUT}\left( \bm{i} \right) \text{\,\,\,\,\,\, for } \Sigma := \bm{i}_1 + \bm{i}_2 + \bm{i}_3,
\]
where $\sigma$ denotes the sum of the carry-in and the two inputs, and $\mathbbm{1}_{\text{condition}}$ is the binary indicator function yielding $1$ if the condition is satisfied and $0$ otherwise.

\begin{figure}[t!]
    \centering
    \scalebox{0.95}{
        \includegraphics[width=\columnwidth]{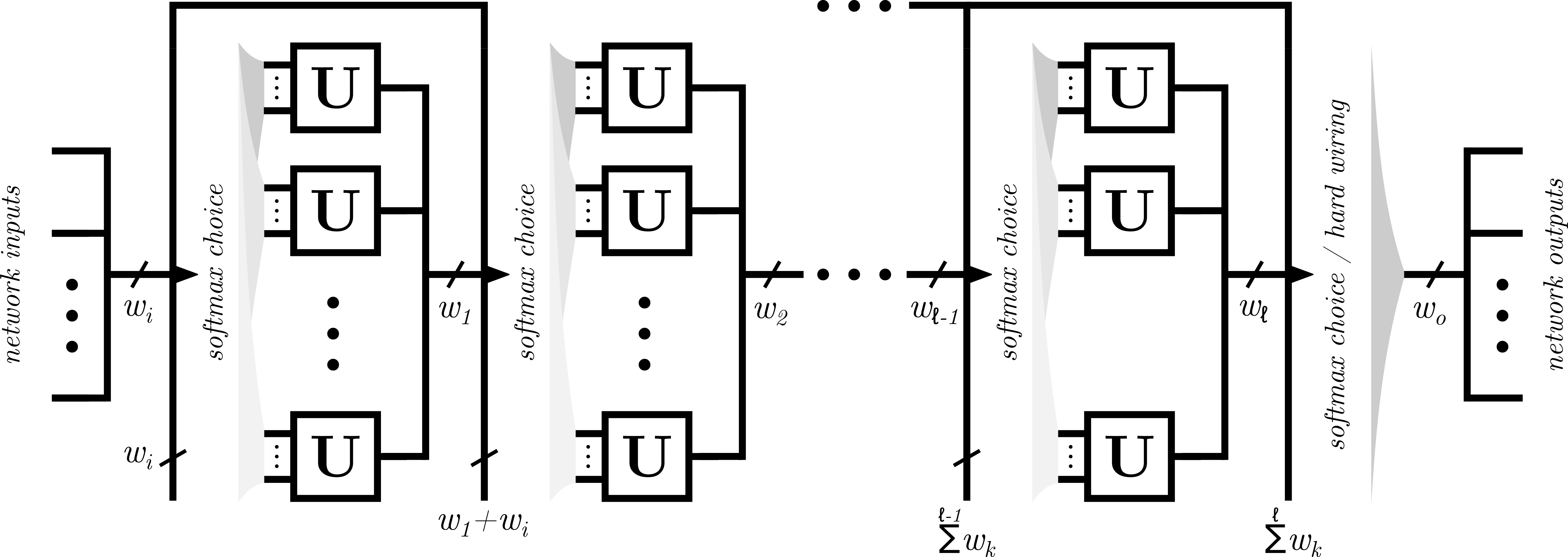}
    }
    \caption{
         An illustration of the universal unit arrangement (\textbf{U}) and softmax-choice wiring as described in \Cref{section:softax-choice_wiring}.
         Mimicking residual connections from convolutional neural networks, the outputs of all previous layers of universal units are combined and then offered for connection to the current layer. The output of the last layer is either hardwired or selected by constant attention to yield the outputs of the network.
    }
    \label{figure:wiring}
\end{figure}

\subsection{Softmax-Choice Wiring}
\label{section:softax-choice_wiring}
We stack universal units in layers $1 \leq k \leq \ell$ of widths $w_k$.
For simplicity, we refer to the inputs of the network as the $0$th layer and have $w_0 := w_i$.
Let $\bm{i}^{k,m}_p$ be the $p$th input of the $m$th unit in layer $k \geq 1$ where $1 \leq m \leq w_k$, and let $o^{k,m}$ be the output of the $m$th unit in layer $k \geq 0$.
Then, for each $k \geq 1$ and $m,p$ as above, the $p$th input of the $m$th unit in the $k$th layer is computed as
\[
    \bm{i}_p^{k,m} := \sum\limits_{l = 0}^{k-1} \sum\limits_{n = 1}^{w_l} o^{l,n} \textit{softmax}\left( \bm{c}^{k,m,p} \right)_{l,n},
\]
where $\bm{c}^{k,m,p}$ is a learned bias matrix of dimension $k \times \max_{0 \leq l < k}{w_l}$ and the softmax is computed over both dimensions.

In other words, every input to a unit is computed by applying constant attention to the outputs of all the previous layers,  input layer included.
Finally, outputs of the last layer are designated as the outputs of the network.
This can be done by hard-wiring a selection of unit outputs to the $w_o$ network outputs, or by simply adding one more layer of $w_o$ softmax choices.
The entire wiring approach is depicted in \Cref{figure:wiring}.

\subsection{Output Loss and Sharpening Softmax Loss}
We use per-signal binary cross entropy loss to compute the loss on the outputs of the network.
Further, since we implicitly expect every softmax signal selection (both within units and wiring between units) to eventually rely only on a single output, we added a sharpening entropy loss, controlled by a hyperparameter $\sigma$.
More specifically, for an $n$-dimensional softmax choice $\bm{s}$, we compute
$
    H\left(\bm{s}\right) = \sum_{m = 1}^n - \bm{s}_m \log \bm{s}_m.
$
It is the sharpening loss that forces all softmax choices to eventually become $0$ or $1$, warranting explainability when inspecting the networks after the training concludes.
The total training loss for true example outputs $\bm{o}_\text{T}$ given prediction outputs $\bm{o}_\text{P}$ is then
$
    \textit{loss}\left(\bm{o}_\text{T}, \bm{o}_\text{P}\right) := \text{BCE}\left(\bm{o}_\text{T}, \bm{o}_\text{P}\right) + \sigma\sum_{\bm{s} \in S} H\left(\bm{s}\right),
$
where $S$ is the set of all softmax choices made within the network.
Once the training has concluded, the synthesised network can be read out using the procedure described in \Cref{appendix:explainability}.

\section{Evaluation}
\label{section:evaluation}
We evaluate our method on datasets comprising input-output examples for a number of tasks, namely: arithmetic negation, subtraction, addition, multiplication, long division, remainder computation, bitwise and, or, xor, not, bit shifts left and right, line multiplexing, de-multiplexing, decoding, and priority encoding.
These were inspired by \cite{rokach2012machine}.
Each task is described in detail in \Cref{appendix:tasks}.

Two training datasets, \texttt{EC-2-100} and \texttt{EC-4-100}, are formed from the examples for the above tasks where the key inputs or outputs are 2 bits and 4 bits wide, respectively.
Four more testing datasets are then generated by leaving out 5\% and 10\% of the examples for each training dataset.
Thus, we consider six datasets in total, four of which test the ability of the given network to generalise already-seen behaviour to previously unseen inputs.
We denote these datasets by strings of the form \texttt{EC-width-ccc}, where \texttt{EC} stands for ``elementary circuits'', \texttt{w} denotes width and is \texttt{2} or \texttt{4}, and \texttt{ccc} (denoting completeness) is \texttt{100} (all examples), \texttt{95} (5\% dropped out), or \texttt{90}.

We find that the choice of the universal unit significantly influences the accuracy of the method.
AIG and LUT networks have the best and worst accuracy scores, respectively.
Moreover and perhaps surprisingly, the best-performing LUT configurations did not respond to the changes to the proportion of training examples being left out.
This suggests that the LUT networks may be implicitly internalising the desired functionality in the loss minimisation process, relying on the broader picture of the behaviour rather than individual example pairs.

LUT-adder blocks are the only type of universal unit we considered with explicitly incorporated bias towards the tasks.
We observe that their performance is only slightly lower than that of LUTs, likely due to the higher number of learnable parameters, but that it increases with the \textit{decreasing} number of training examples.
This can be interpreted as the LABs finding it easier to learn inductive behaviour than its peers, and easier against its own performance when only few key examples are provided.

In sum, the performance of LUTs suggests that even neural networks tailored to a specific purpose show signs of developing an high-level picture of the task, while the performance of LABs shows the utility of architecturally incorporating implicit knowledge about the task at hand for its inductive performance on discrete, logical tasks.

\begin{table*}[b!]
\centering
\scalebox{1.00}{
    \begin{tabular}{l|c|c|c|c}
    
    \toprule
    Dataset & Metric Type  & \multicolumn{3}{c}{Universal Unit Type} \\
    \midrule
    
    \multicolumn{2}{c}{}  & \multicolumn{1}{c|}{2-bit AIG} & \multicolumn{1}{c|}{4-bit LUT} & \multicolumn{1}{c}{4-bit LAB} \\

    \midrule
        \multirow{2}{*}{\textsc{EC}-$2$-$100$}      & signal  &     \, \bw{0.956} \,&\,	\bw{\bo{0.969}}  \,&\,    0.961     \\
                                                    & example &     \, \bw{0.930} \,&\, \bw{\bo{0.953}}  \,&\,	 0.938      \\
        
        \multirow{2}{*}{\textsc{EC}-$2$-$95$}       & signal  &     \, 0.956 \,&\,	\bw{\bo{0.969}}  \,&\,	 0.964          \\
                                                    & example &     \, 0.922 \,&\,	\bw{\bo{0.953}}  \,&\,	 0.941          \\
        
        \multirow{2}{*}{\textsc{EC}-$2$-$90$}       & signal  &     \, 0.941 \,&\,	\bw{\bo{0.969}}  \,&\,	\bw{\bo{0.969}} \\
                                                    & example &     \, 0.898 \,&\,	\bw{\bo{0.953}}  \,&\,	\bw{\bo{0.969}} \\
    
     \midrule
                                                        
         \multirow{2}{*}{\textsc{EC}-$4$-$100$}     & signal  &    \, \bw{0.925} \,&\,	\bw{\bo{0.967}}  \,&\,	0.952       \\
                                                    & example &     \, \bw{0.778} \,&\,	\bw{\bo{0.889}}  \,&\,	0.849       \\
        
        \multirow{2}{*}{\textsc{EC}-$4$-$95$}       & signal  &     \, 0.923 \,&\, \bw{\bo{0.967}}  \,&\,	0.960           \\
                                                    & example &     \, 0.735 \,&\,	\bw{\bo{0.889}}  \,&\,	0.868           \\
        
        \multirow{2}{*}{\textsc{EC}-$4$-$90$}       & signal  &     \, 0.922 \,&\,	\bw{\bo{0.967}}  \,&\,	\bw{\bo{0.967}} \\
                                                    & example &     \, 0.723 \,&\,	\bw{\bo{0.889}}  \,&\,	\bw{\bo{0.889}} \\

    \bottomrule 
    
    \end{tabular}
}
\caption{
    The results of a systematic evaluation of our method on the datasets.
    \bo{Emphasis} and \bw{emphasis} mark the best results per dataset and per-unit, respectively.
    Each experiment was run with 20 different configurations but training and initialisation seeds fixed, as described in \Cref{appendix:hyperparameters}.
}
\label{table:summary_results}
\end{table*}

\newpage
\bibliographystyle{plain}
\bibliography{bibliography}

\newpage
\section*{Checklist}

\begin{enumerate}

\item For all authors...
\begin{enumerate}
  \item Do the main claims made in the abstract and introduction accurately reflect the paper's contributions and scope?
    \answerYes{}
  \item Did you describe the limitations of your work?
    \answerYes{}
  \item Did you discuss any potential negative societal impacts of your work?
    \answerNA{}
  \item Have you read the ethics review guidelines and ensured that your paper conforms to them?
    \answerYes{}
\end{enumerate}

\item If you are including theoretical results...
\begin{enumerate}
  \item Did you state the full set of assumptions of all theoretical results?
    \answerNA{}
        \item Did you include complete proofs of all theoretical results?
    \answerNA{}{}
\end{enumerate}

\item If you ran experiments...
\begin{enumerate}
  \item Did you include the code, data, and instructions needed to reproduce the main experimental results (either in the supplemental material or as a URL)?
    \answerYes{}{}
  \item Did you specify all the training details (e.g., data splits, hyperparameters, how they were chosen)?
    \answerYes{Please see \Cref{appendix:hyperparameters}.}
        \item Did you report error bars (e.g., with respect to the random seed after running experiments multiple times)?
    \answerNA{}
        \item Did you include the total amount of compute and the type of resources used (e.g., type of GPUs, internal cluster, or cloud provider)?
    \answerYes{Please see \Cref{appendix:hyperparameters}.}
\end{enumerate}

\item If you are using existing assets (e.g., code, data, models) or curating/releasing new assets...
\begin{enumerate}
  \item If your work uses existing assets, did you cite the creators?
    \answerNA{}
  \item Did you mention the license of the assets?
    \answerNA{}
  \item Did you include any new assets either in the supplemental material or as a URL?
    \answerNA{}
  \item Did you discuss whether and how consent was obtained from people whose data you're using/curating?
    \answerNA{}
  \item Did you discuss whether the data you are using/curating contains personally identifiable information or offensive content?
    \answerNA{}
\end{enumerate}

\item If you used crowdsourcing or conducted research with human subjects...
\begin{enumerate}
  \item Did you include the full text of instructions given to participants and screenshots, if applicable?
    \answerNA{}
  \item Did you describe any potential participant risks, with links to Institutional Review Board (IRB) approvals, if applicable?
    \answerNA{}
  \item Did you include the estimated hourly wage paid to participants and the total amount spent on participant compensation?
    \answerNA{}
    
\end{enumerate}

\end{enumerate}

\appendix

\newpage
\section{Tasks of the \texttt{EC-w-100} Datasets}
\label{appendix:tasks}

A number of bitwise, arithmetic, and signal control learning tasks form the complete datasets of width \texttt{w}.
We refer to the individual training instances as tasks, since they specify behaviour that is to be learned from input-output examples.
The behaviour of each task is given as a sequence of $\left(x, y\right)$ pairs, where $x$ is a binary vector that is to be fed into logic circuit, and $y$ is the binary output representing the desired output for input $x$.
The dimensions of vectors $x,y$ depend on the task and on \texttt{w}.

\subsection{Bitwise Operations}
\subsubsection{Bit Negation}
The dimension of both $x$ and $y$ is exactly \texttt{w}.
The input-output pairs for this task give the behaviour of a \texttt{w}-bit logical NOT gate.
For example, for $x = 1011_2, y = 0100_2$.

\subsection{Bitwise AND, OR, and XOR}
The dimension of $x$ is $2\texttt{w}$, the dimension of $y$ is \texttt{w}.
The input-output pairs describe the behaviour of the \texttt{w}-bit AND/OR/XOR gate, performing the operation on the two \texttt{w}-bit slices of $x$.
For example, $\text{AND}\left(1011_20110_2\right) = 0010_2$.

\subsection{Unsigned Bit Shift to the Left/Right}
The dimension of both $x$ and $y$ is exactly \texttt{w}.
The input-output pairs are given by shifting $x$ to the left/right by one bit and filling the empty bit with $0$.
As an example, $\text{SHL}\left(1011_2\right) = 0110_2$.

\subsection{Arithmetic Operations}
\subsubsection{Arithmetic Negation}
The dimension of both $x$ and $y$ is exactly \texttt{w}.
The task is to perform arithmetic negation (in the two's complement system for bit width \texttt{w}).
This is equivalent to performing the ones' complement and then adding $1$, e.g. $\text{NET}\left(1110_2\right) = 0010_2$.

\subsubsection{Arithmetic Addition and Subtraction}
The dimension of $x$ is $2\texttt{w}$, the dimension of $y$ is $\texttt{w}+1$.
The task is to perform arithmetic addition/negation, with negative numbers represented in the two's complement system of widths \texttt{w}, $\texttt{w}+1$ for inputs and outputs, respectively.
The $\texttt{w}+1$st bit is the carry/borrow bit, respectively.
As an example, $\text{SUB}\left(0001_21110_2\right) = 00011_2$.

\subsubsection{Arithmetic Multiplication}
Both $x$ and $y$ are of dimension $2\texttt{w}$.
Unsigned arithmetic multiplication is performed on the individual slices of $x$.
For example, $\text{MUL}\left(0011_21110_2\right) = 00101010_2$ because $3 \times 14 = 42$.

\subsubsection{Integral Division and Remainder (Modulo)}
The dimension of $x$ is $2\texttt{w}$, the dimension of $y$ is \texttt{w}.
The output is the output of unsigned long division (remainder) of the value of the first \texttt{w}-bit slice of $x$ by the value of the second.
For example $\text{DIV}\left(0110_20011_2\right) = 0011_2, \text{REM}\left(0110_20011_2\right) = 0001_2$.

\subsection{Signal Control}
\subsubsection{$\texttt{w}$-to-$1$ Multiplexer}
The dimension of $x$ is $\texttt{w} + \floor{\log_2{\texttt{w}}}$, the dimension of $y$ is $1$.
The first \texttt{w}-bit slice of $x$ contains the multiplexer's input signals.
The remaining slice of $x$ contains the value deciding which of the signals to output, with the lines being numbered from $0$.
For an example with $w = 5$, to choose the 3rd signal from $01101_2$, $\text{MUX}\left(0111_211_2\right) = 0_2$.
The value of the binary logarithm is rounded to avoid having to specify undefined behaviour.

\subsubsection{$1$-to-\texttt{w} Demultiplexer}
The dimension of $x$ is $1+\floor{\log_2{\texttt{w}}}$, the dimension of $y$ is $\texttt{w}$.
The value of the slice of $x$ that runs from the second bit onwards describes which output line (numbering from $0$) to forward the input signal (stored in the first bit of $x$) to.
The remaining output lines are set to $0$.
As an example for $w = 4$, $\text{DEMUX}\left(1_211_2\right) = 1000_2$.

\subsubsection{$\floor{\log_2{\texttt{w}}}$-to-$\texttt{w}$ Line Decoder}
The dimension of $y$ is \texttt{w}, the dimension of $x$ is $\floor{\log_2{\texttt{w}}}$.
The line decoder drives the $x$th output line to $1$ and all the remaining lines to $0$, as in $DEC\left(10_2\right) = 0100_2$ for $\texttt{w} = 4$.

\subsubsection{$\texttt{w}$-to-$\ceil{\log_2{\texttt{w}}}$ Priority Encoder}
The dimension of $x$ is \texttt{w}, the dimension of $x$ is $\ceil{\log_2{\texttt{w}}}$.
The priority encoder takes the \texttt{w} signals of $x$ as inputs and outputs binary value representing the position (numbered from $0$) of the first non-zero line on its input.
For an example with $w = 5$, $\text{ENC}\left( 00110_2 \right) = 001_2$.

\newpage
\section{Explainability -- Reading Out the Learned Network Logic}
\label{appendix:explainability}
Once the training has concluded and the sharpening loss has converged, one can proceed to read out the synthesised network logic as described below.

\textbf{Initialisation.} If the selector had been used, its softmax choices (now sharp -- either $0$ or $1$ per signal line) are used to find the output units $\mathcal{O}$.
If it had not been used, the $\mathcal{O}$ is set to the hard-wired output units.

\textbf{Wire identification.} In line with the notation and definitions of \Cref{section:softax-choice_wiring}, define the wire presence indicator $\omega\left(U, U', p'\right)$ for $U,U'$ units of the network used for training as
\[
    \omega\left( U^{k,m}, U^{k', m'}, p' \right) := 
    \begin{cases} 
      1 & \text{ if } \textit{softmax}\left( \bm{c}^{k',m',p'} \right)_{k,m} > \tau \\
      0 & \text{ otherwise}
   \end{cases},
\]
where $\tau \in \left(0,1\right)$ is a threshold for the softmax values for which the wire is to be considered present.
In our experimentation, we found that the appropriate threshold depends on the values of $\sigma$, learning rate, and the number of epochs (``tightness'' of convergence), but could usually comfortably be $0.95$ or above.

Put in words, $\omega\left(U_1,U_2,p\right) = 1$ means that the output of $U_1$ is connected to the $p$-th input of $U_2$ in the synthesised network.

\textbf{Network Extraction.} With $\omega$ defined as above, the units $\mathcal{U}$ used by the synthesised network can be extracted recursively as follows:
\begin{align*}
    \mathcal{U}_0 & := \mathcal{O} \\
    \mathcal{U}_i & := \left\{
        U^{k,m} : \exists k',m',p'\exists k < k' \exists 1 \leq m \leq w_{k} \text{ s.t. } \omega\left( U^{k,m}, U^{k', m'}, p' \right) = 1
    \right\} \text{ for } 1 \leq i \leq \ell + 1 \\
    \mathcal{U} & := \bigcup\limits_{i = 0}^{\ell + 1} \mathcal{U}_i
\end{align*}

Then the pair $\left(\mathcal{U}, \omega\right)$ gives the synthesised network.

\newpage
\section{Training and Testing Configurations}
\label{appendix:hyperparameters}

\subsection{Train-Test Splits}
Given the nature of the tasks considered, the training and test dataset are largely the same, depending on the situation.

For \texttt{EC-2-100} and \texttt{EC-4-100}, the two sets are identical.
For all other datasets \texttt{EC-w-cc}, the networks are trained on \texttt{EC-w-100} and \texttt{EC-w-100} and then tested on a fixed subset \texttt{EC-w-cc}, where is the percentual proportion of the examples of \texttt{EC-w-100} that were retained for testing.

\subsection{Architecure}
For each unit and each task in each dataset, we considered a maximum of 4 layers of 40 units for the network to emerge through softmax choices.
Half of our configurations further added softmax selections for the outputs of the final layer as per \Cref{section:softax-choice_wiring}.

Overall, we observed improvements in performance for AIG networks that had the softmax output selectors but performance deterioriation for LUT and LAB networks in the corresponding configurations.

\subsection{Initialisation and Data Shuffle Seeds}
These were fixed, with the particular choices detailed in the code.

\subsection{Training Parameters and Hyperparameters}
A total of 20 configurations was considered for the test runs, and each run was run for a maximum of 100 epochs with early stopping for when the perfect accuracy of $1$ has been achieved.

The fixed batch sizes lied in $\left[4, 16\right]$, learning rates in $\left(0.1, 0.6\right)$, learning rate exponential decay factors in $\left(0.9, 1.0\right)$.
We used the Adam optimiser.

After initial experimentation, we fixed $\sigma = 1.0$, but engaged it only in the second half of the training (i.e. after 50 epochs).

\subsection{Computational Resources}
Each of our experiments was run on a single core of a cluster of Dual Deca-Core Intel Xeon E5-2690 v2 processors and fit within $10$ GiB of RAM memory.
In this setup, all experiments terminated within 12 hours.

\end{document}